\pdfoutput=1
\documentclass{article}
\usepackage{ijcai22}
\usepackage{times}
\usepackage{soul}
\usepackage{url}
\usepackage[hidelinks]{hyperref}
\usepackage[utf8]{inputenc}
\usepackage[small]{caption}
\usepackage{graphicx}
\usepackage{amsmath}
\usepackage{amsthm}
\usepackage{booktabs}
\usepackage{algorithm}
\usepackage{algorithmic}
\usepackage{here}
\usepackage{ifthen}
\usepackage{bibunits}
\usepackage{comment}
\begin{document}


\newcounter{language}
\setcounter{language}{1}


\newcommand{\proposalfirst}{ChAt, Shift and PERform}
\newcommand{\proposal}{Chat, Shift and Perform}
\newcommand{\PROP}{CASPER}
\newcommand{\lone}{chatter}
\newcommand{\ltwo}{shifter}
\newcommand{\lthree}{performer}
\newcommand{\Lone}{Chatter}
\newcommand{\Ltwo}{Shifter}
\newcommand{\Lthree}{Performer}

\title{\proposal: Bridging the Gap between Task-oriented and Non-task-oriented Dialog Systems}

\author{
  Teppei Yoshino$^1$
  \and
  Yosuke Fukuchi$^1$\and
  Shoya Matsumori$^1$\And
  Michita Imai$^1$
  \affiliations
  $^1$Keio University, Yokohama, Japan\\
  \emails
  tepei@keio.jp,
  \{fukuchi, shoya, michita\}@ailab.ics.keio.ac.jp
}

\maketitle
\begin{abstract}
  We propose CASPER (ChAt, Shift and PERform), a novel dialog system consisting of three types of dialog models: \textit{chatter}, \textit{shifter}, and \textit{performer}. Shifter, which is designed for topic switching, enables a seamless flow of dialog from open-domain chat- to task-oriented dialog.
  In a user study, CASPER gave a better impression in terms of naturalness of response, lack of forced topic switching, and satisfaction compared with a baseline dialog system trained in an end-to-end manner. In an ablation study, we found that naturalness of response, dialog satisfaction, and task-elicitation rate improved compared with when shifter was removed from CASPER, indicating that topic shift with shifter supports the introduction of natural task-oriented dialog.
\end{abstract}
\section{Introduction}
With the increase in the use of smart speakers, dialog systems have become more common in daily life. Unlike store-assistant robots with the main purpose of selling products in stores, these dialog systems are used for general purposes and are expected to perform a variety of task-oriented dialogs, including system-driven tasks such as item recommendation.
In addition, users expect to be able to chat with such systems. At least 38\% of calls to a dialog system were chatting~\cite{chat_detection}.
Therefore, dialog systems should support both chat- and task-oriented dialog.
\begin{figure}[t]
  \centering
  \includegraphics[width=\linewidth,bb=0 0 395 323,clip,page=1]{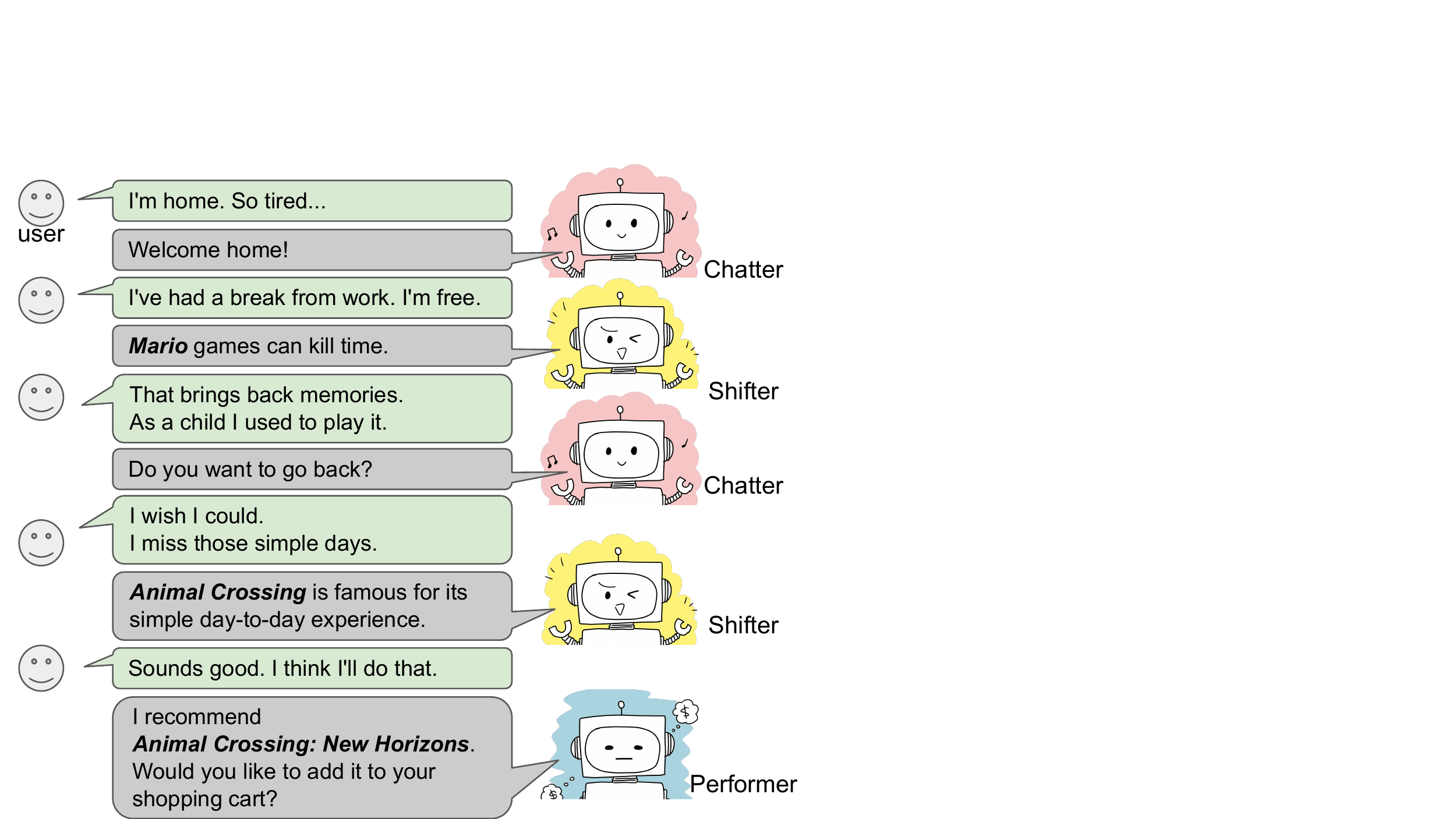}
  \caption{\textbf{Example of dialog between user and {\PROP}}.
   Chatter conducts chatting, shifters change topics from chatting to the target domain of the system's tasks, and performers perform tasks.
  }
  \label{fig:eyecatch}
  \vspace{-1em}
\end{figure}

There are two main types of dialog systems that manage both chat- and task-oriented dialog: single-model and bi-model.
A single-model dialog system executes both chat- and task-oriented dialog using one model. A representative single-model system uses the knowledge-grounded Seq2Seq model~\cite{fb_end2end,akgcm,ner_seq2seq,cr_walker}.
 A bi-model system combines a chat-oriented dialog model with a task-oriented dialog model~\cite{multi_expert_model,chat_detection,adversarial_chat_detection}. \cite{multi_expert_model} implemented both chat- and task-oriented dialog models in a single robot. In addition, models proposed by~\cite{chat_detection,adversarial_chat_detection} determine whether dialog systems should respond in the chat- or task-oriented domains according to a user utterance by detecting whether the utterance is for chat- or task-oriented dialog.

Although these studies focused on both chat and task-oriented dialog, few studies have focused on the boundary between two types of dialog. Specifically, single-model dialog systems produce only a likely utterance based on a training dataset and cannot control the flow of dialog between chat- and task-oriented dialog. Such systems sometimes continue chatting and cannot move on to a task. (1) Bi-model dialog systems do not take into account an \textit{intermediate utterance} between chat- and task-oriented dialog, such as an utterance to guide the topic to a specific item that the system wants to recommend, so they cannot provide seamless transition. In addition, the switching between chat- and task-oriented dialog in prior studies was reactive, and (2) it is still unclear when to switch for a system to actively influence the dialog. Therefore, with current dialog systems,  task-oriented dialog is suddenly initiated in the midst of chatting, resulting in a breakdown in the dialog.
To address these issues, a dialog system should first allow for both chat- and task-oriented dialog to respond to open-domain chatting then gradually change the topic for carrying out a task.

We propose \textit{{\PROP}: {\proposalfirst}} a dialog system consisting of three types of dialog models ({\lone}, {\ltwo}, and {\lthree}) as well as a selector. {\PROP} was designed to seamlessly switch the dialog domain from chat- to task-oriented dialog to accomplish a task (Fig.\ref{fig:eyecatch}).
Chatter and performer are equivalent to bi-model chat- and task-oriented dialog systems, respectively. The unique features of our system are {\ltwo} and selector.
To tackle the problem (1) and make smooth topic transition possible, {\ltwo} was designed to generate an intermediate utterance. By training on a dataset of dialog that ends with responses related to the system's task domain (target domain), shifter generates intermediate utterances that seamlessly bridges gap between chat- and task-oriented dialog.
In addition, selector determines when to activate the appropriate model for the problem (2).

In a user study, participants were asked to answer a questionnaire regarding their impressions of interaction with CASPER. As a result, CASPER was significantly better in terms of naturalness of responses and smoothness of topic transitions compared with single- and bi-model systems.

\section{Related Work}
\noindent\textbf{Task-oriented dialog systems.}
  The task-oriented dialog systems that have been extensively studied are conversational recommender systems (CRSs).~\cite{crm} is a typical CRS that uses natural language understanding to track the user's beliefs and implements reinforcement learning to determine the policy of questions and recommendations.
However, most of the existing task-oriented dialog systems do not assume out-of-domain (OOD) utterance by nature.

\noindent\textbf{Non-task-oriented dialog systems.}
  The most basic non-task-oriented dialog systems are based on the Seq2Seq model~\cite{seq2seq} that consists of encoder decoder pairs.
  When the Seq2Seq model is trained using maximum likelihood estimation (MLE),
  it generates generic responses such as ``\textit{I don't know}'' because it is 
  assumed that with MLE that there is a 1-to-1 context-response relationship, resulting in a dialog that does not change the topic. This is a problem when considering the control from chat- to task-oriented dialog. To address this issue, several studies have investigated generating distinctive responses.~\cite{less_generic_response} solved this problem by applying the objective function to assume a one-to-many problem for natural language generation (NLG).~\cite{emo_hred} controls the training of NLG by biasing the training corpus beforehand.

In our {\PROP}, a chat-oriented dialog system, {\lone} continues the dialog with simple responses, while {\ltwo} generates more meaningful responses to develop the conversation, thus solving the problem of generic response. The {\ltwo}'s response generation can be controlled with biasing a corpus for training~\cite{emo_hred}.

\noindent\textbf{Hybrid model for task- and non-task-oriented dialog systems.}
  As mentioned above, there are two main types of dialog systems for handling both task- and non-task oriented dialog: single-model that for both types of dialog using one model, and bi-model approach that uses two types of models.
  Representative single-model dialog systems use the knowledge-grounded Seq2Seq model~\cite{fb_end2end,akgcm,ner_seq2seq,cr_walker}. These systems use Seq2Seq both for generating chatting responses and carrying out tasks. 
  For example,~\cite{fb_end2end} acquires task-specific knowledge by pre-training in an end-to-end manner. The system proposed by~\cite{akgcm} extracts knowledge from outside the language model.~\cite{ner_seq2seq} applied named entity recognition (NER) before input so that the model does not need to handle task-specific queries. In addition,~\cite{cr_walker} learns language processing and recommendation-entity selection in an end-to-end manner.

  Bi-model dialog systems that combine the chat- and task-oriented dialog models are being actively studied. For example, the multi-expert model~\cite{multi_expert_model} adds a chat task to a robot capable of performing various tasks and implements both chat- and task-oriented dialog in a single framework. \cite{chat_detection,adversarial_chat_detection} detects whether the query is for chat- or task-oriented dialog, enabling bi-model dialog systems to determine whether they should respond in the chat- or task-oriented domains.

Although these studies contributed to the compatibility of chat- and task-oriented dialog, there has not been sufficient research on the boundary between these types of dialogs. Specifically, single-model dialog systems do not control the switching between the two types of dialog, while the bi-model dialog systems have a gap between them and do not provide a seamless connection.
With single-model systems, it is difficult to control the flow of dialog domain, since the model only follows the scenario of the training corpus.
Bi-model systems are supposed to respond with chat for chat queries and task execution for task-specific queries, and not supposed to seamlessly switch between chat- and task-oriented dialog.
If dialog system that combines both chat- and task-oriented dialog cannot handle topic transitions between the two, the following problems may arise.
(a) The chatting just continues and the task is not accomplished.
(b) Task-oriented dialog is suddenly initiated in the midst of chatting, resulting in a breakdown in the dialog.
To address these issues, a dialog system should allow for both chat- and task-oriented dialog to first respond to open-domain chat then gradually change the topic for carrying out a task.

{\PROP} solves the topic transition problem by training {\ltwo} with a dialog dataset in which topic transition occurs. The timing of the topic transition is determined by likelihood estimation using the transition history of the dialog.

\begin{figure*}[t]
  \centering
  \includegraphics[width=\linewidth,bb=0 0 520 112,clip]{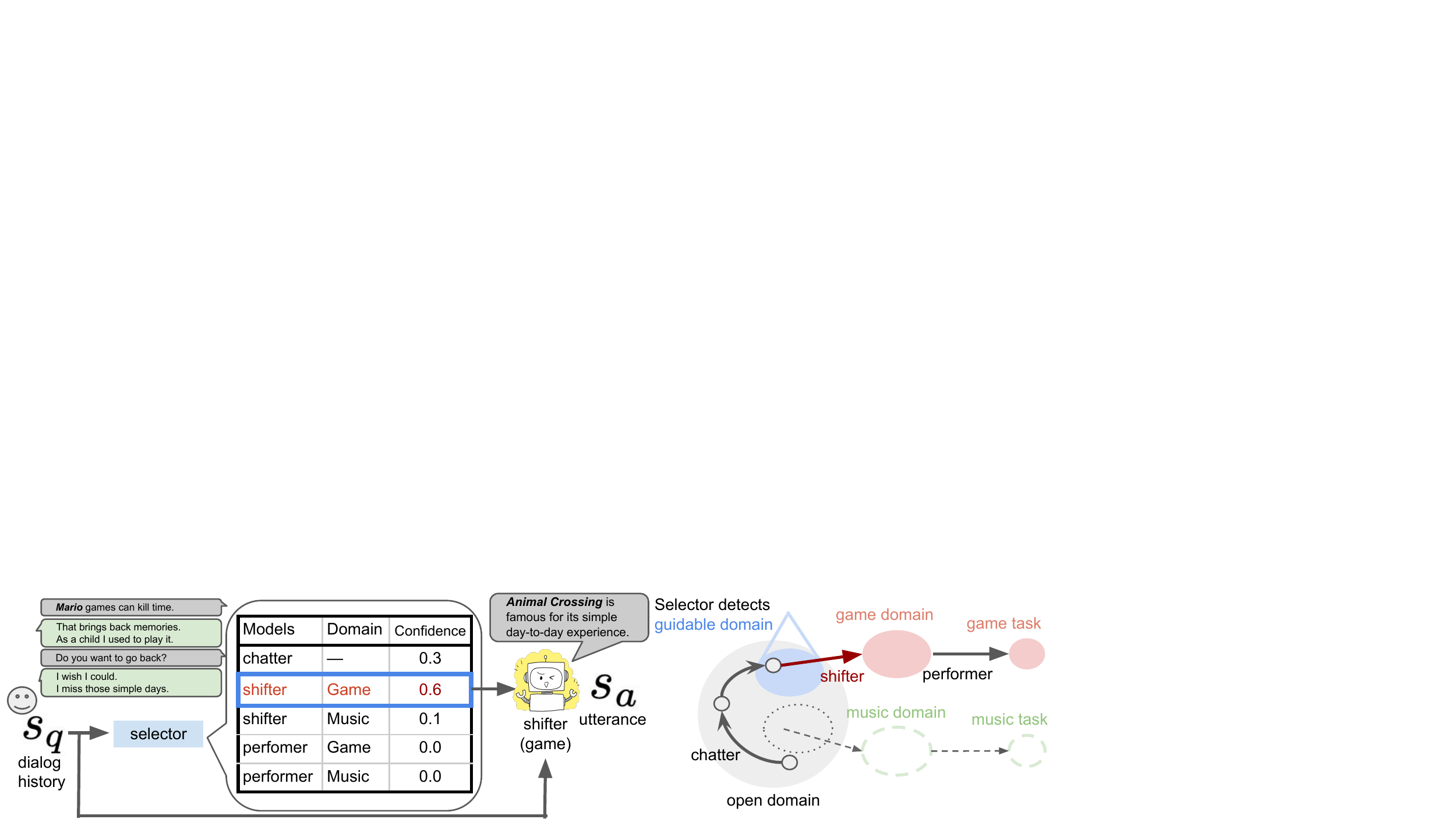}
  \caption{\textbf{The whole architecture of {\PROP}}.
  CASPER consists of three types of dialog systems: chatter, shifter, and performer, and a selector that decides the response model among them.
  }
  \label{fig:system}
    \vspace{-1em}
\end{figure*}

\section{\proposal}

\subsection{General design of CASPER}
\textit{{\PROP}} steers the dialog domain from chat- to task-oriented dialog without breaking down the dialog by using {\ltwo} to generate topic-changing utterances and selector to determine the timing and destination of topic transitions (Fig.\ref{fig:system}).

As mentioned above, CASPER consists of three types of dialog models: chatter, shifter, and performer, as well as selector. Chatter is a chat-oriented dialog model that conducts chatting. Shifter guides topics from chatting to a pre-defined task domain. Performer performs task-oriented dialog after the topic change. CASPER can have multiple performers, and each performer has its corresponding shifter. Selector is responsible for determining when and what model to activate in the dialog.
The input of chatter, shifter, and performer is the dialog history $s_q$, and the output is the response sentence $s_a$.
Shifter solves the problem with prior studies on bi-model dialog systems of not being able to handle topic transition dialog between chat- and task-oriented dialog.
Selector solves the problem of uncontrollable response strategies due to the corpus scenario, which has been an issue in prior studies on single-model dialog systems.

\begin{table*}[t]
  \centering
  \begin{tabular}{rlcc}
    \hline
    Label        & Text                                               & Negative label        & Positive label     \\
    \hline\hline
        Naturalness  & Was the response of this dialog system natural to your utterance?     & very unnatural     & very natural     \\
        Attachment   & Would you like to have this AI in your home?                & not at all & definitely want \\
        Satisfaction & How satisfied are you with the conversation?                         & not satisfied       & very satisfied     \\
        Transition   & Was the dialog system's choice of topic forceful or natural?          & very forced   & very natural     \\
        Favor        & Did you like the products recommended by the dialog system? & hated them & loved them     \\
    \hline

  \end{tabular}
  \caption{
    \label{table:eval_items}\textbf{Questionnaire items regarding conversational recommendation}.
    Participants responded to each item on 7-point scale, with 1 being negative label and 7 being positive label.}
\end{table*}

\subsection{Chatter and performer}
Chatter is an open-domain and chat-oriented dialog model that can be implemented using current neural dialog models.

Performer is a task-oriented dialog model that can be implemented using prior research on task-oriented dialog systems, such as rule-based systems for simple single-turn question answering, or more complex multi-turn dialog systems such as a CRS.
\subsection{{\Ltwo}}
Shifter is a dialog model that generates responses that bridge chatter and performer.
Shifter is expected to generate an utterance that seamlessly guides the dialog topic from chat to a specific task domain performer is in charge of.
For training shifter, we collected context-reply pairs that included topic transitions to a specific domain. To generate utterances that change the topic from outside to inside the target domain, we performed transfer learning on the chatter in accordance with the following loss function:
\begin{align}
  L(s_q, s_a) & = -\sum_{t=1}^T \log p(w_t^a\ |\ s_q, w_{<t}^a)\label{equaton: shifter} \\
  s_q         & =\left\{w^q \ |\ \forall w^q \notin D\right\}\notag                     \\
  s_a         & =\left\{w^a \ |\ \exists w^a \in D\right\}\notag,
\end{align}
where  $w_t^a$ is the $t$-th word in the output sentence, $T$ is the number of words in the output sentence, $w^q$ is the word in the input dialog, and $D$ is the set of words belonging to the target domain.
Shifter was trained as a separate model for each domain. This enables it to learn topic transitions unique to each domain, including the use of proper nouns in the domain, as shown in Fig. \ref{fig:eyecatch}.
As we discuss in section \ref{sec:experiment}, a single shifter were to learn transitions to multiple domains, it would converge to a response that is common to all topic transitions to all domains as the number of domains increases, which may result in a general response similar to that of chat-oriented dialog systems. Since the destination domain is indirectly determined by the dialog system, we may not be able to control the destination of the domain.

Shifter is expected to generate an utterance that seamlessly guides the dialog topic from chat to a specific task domain performer is in charge of.
As mentioned above, shifter is trained on a dataset of dialog ending with biased response domain to generate a biased utterance that bridges the gap between chat- and task-oriented dialog.
There is another approach to bias an utterance.~\cite{kbrd} proposed directly manipulating the probability of a token generation in inference step.
In generating shifter's biased response, we did not use a method of directly manipulating the probability of token generation~\cite{kbrd}, but adopted a method of biasing the corpus beforehand~\cite{emo_hred}.
Compared with methods that directly manipulate the output tokens, methods that bias the training corpus are better suited for controlling the general domain but not for controlling utterances at the token level.
Because we use shifter for the rough task of guiding the topic to the target domain, and performer performs the specific dialog about the task, so this is not a problem in our case. In addition, response control by corpus bias has the advantage of maintaining the plausibility of the output sentences since shifter can be trained from the topic transitions that have actually occurred in the corpus.
\subsection{Selector}\label{ss:selector}
Selector is a classifier that determines the response model, given the dialog history.
Each dialog model (chatter, shifter, and performer) has an appropriate context to be activated.
Selector selects when and what model to use to generate response according to the progress of the dialog.

Selector specifies one model as the response model from among chatter and multiple shifter models. However, if performer can respond, it is preferred to the chatter and shifter.
The reason we exclude performer from the response model inference of selector is that task-oriented dialog systems often define their response conditions (e.g., confidence level in FAQ search or keyword matching).

Appropriate switching of the models is expected to enable the introduction of task-oriented dialogs and to improve dialog satisfaction by moderately changing the topic.

Selector is trained by multi-class document classification, which predicts the domain $c_a$ of the future response $s_a$ given the dialog history $s_q$ with the following loss function:
\begin{align}
L(s_q, c_a) = -\sum_i^C c_a^i\log(f(s_q)_i)\label{equation: classification},
\end{align}
where $i$ is the index of the domain assigned to all chatter and shifters in CASPER, $C$ is the number of all domains handled by CASPER including the open-domain in charge of chatter,
$c_a^i$ is 1 if the ground-truth domain of $s_a$ is the $i$-th domain, 0 otherwise. Also, $f(s_q)$ is the output of the feature obtained from $s_q$ into softmax, and $f()_i$ is the confidence that $i$-th is the index of correct answer domain.

In other words, selector is trained by next domain prediction with input of previous dialog history.
Note that we do not use the response sentence $s_a$ to predict its domain $c_a$. The $c_a$ is predicted using only the previous dialog $s_q$ up to the time step before the response $s_a$ is uttered.

Selector uses the confidence of the domain of the next utterance to infer the actual model that should be responded to next in accordance with
\begin{align}
  \newcommand{\argmax}{\mathop{\rm arg~max}\limits}
  \text{Selector}(s_q) = \argmax_{i} \left[\frac{f(s_q)_i}{\alpha_i}\right]\label{equation: selector},
\end{align}
where $\alpha_i$ is a parameter greater than zero that controls how difficult it is to switch to the $i$-th domain. By defining $\alpha_i$, we can easily control the forcefulness of the topic change and the ease of transition to the $i$-th domain. We can also change $\alpha_i$ in time.
In this study, we gradually decreased the alphas for shifters to make them activate more as time passed, so the task could be performed in a less forced way.
When $\alpha$ is the same for all $i$, selector is simply a next-domain estimator that selects the model of the most likely domain.

Document classification for next-domain prediction requires training data of topic-switching dialogs, which is normally expensive to collect. However, in CASPER, we have already collected data on topic switching and non-topic switching in training shifter and chatter, respectively, so we can use these dialog data as training data for the next domain inference. Therefore, for training selector, there is no need to prepare additional data other than the training data of chatter and shifter.

The fact that selector selects a different model at each time step simplifies the role of shifter.
Since shifter performs transfer learning on a corpus consisting of responses of a specific domain, it always returns answers in its responsible domain regardless of the input query. Therefore, if the user continues to interact with shifter alone for a long time, it is expected that the user will get bored and the dialog satisfaction will decrease. However, in CASPER, the user does not continuously interact with a single shifter over a long time step because the model selection of selector occurs every time. Under the assumption that selector selects the response model at each time step, shifter can be used as a response-generation module with the ability to seamlessly pull from various domains into a specific domain.
\begin{table}[t]
  \centering
  \begin{tabular}{rll}
    \hline
                & Baseline  & {\bf{\PROP}}        \\
    \hline\hline
    Perplexity  & \bf{5.94} & 6.16 (avg)          \\
    Naturalness & 3.41      & {\bf4.20} *(p=0.012) \\
    \hline
  \end{tabular}
  \caption{
    \textbf{Results of perplexity and naturalness evaluation}.\label{table:ppl}
    We compared fluency and naturalness of baseline and CASPER using perplexity during training and naturalness ratings.}
\end{table}

\begin{figure}[t]
  \centering
  \includegraphics[width=\linewidth,bb=0 0 450 280,clip,page=1]{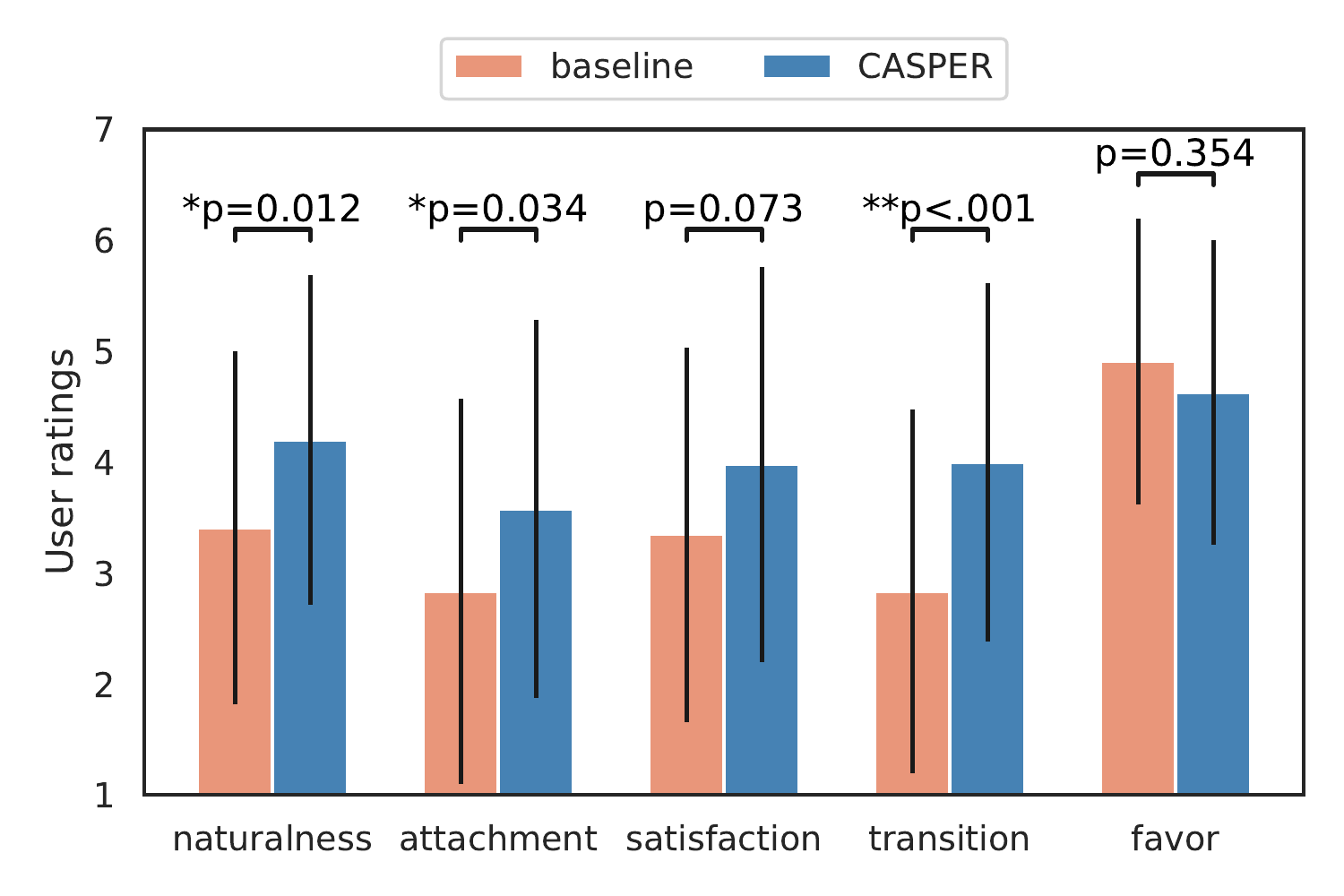}
  \caption{
    \textbf{Comparison of baseline and CASPER user ratings for each question item}.\label{fig:base_casper}
   Mean and standard deviation of the ratings for each item are shown as histograms and bars, respectively.}
\end{figure}

\begin{figure}[t]
  \centering
  \includegraphics[width=\linewidth,bb=0 0 430 280,clip,page=1]{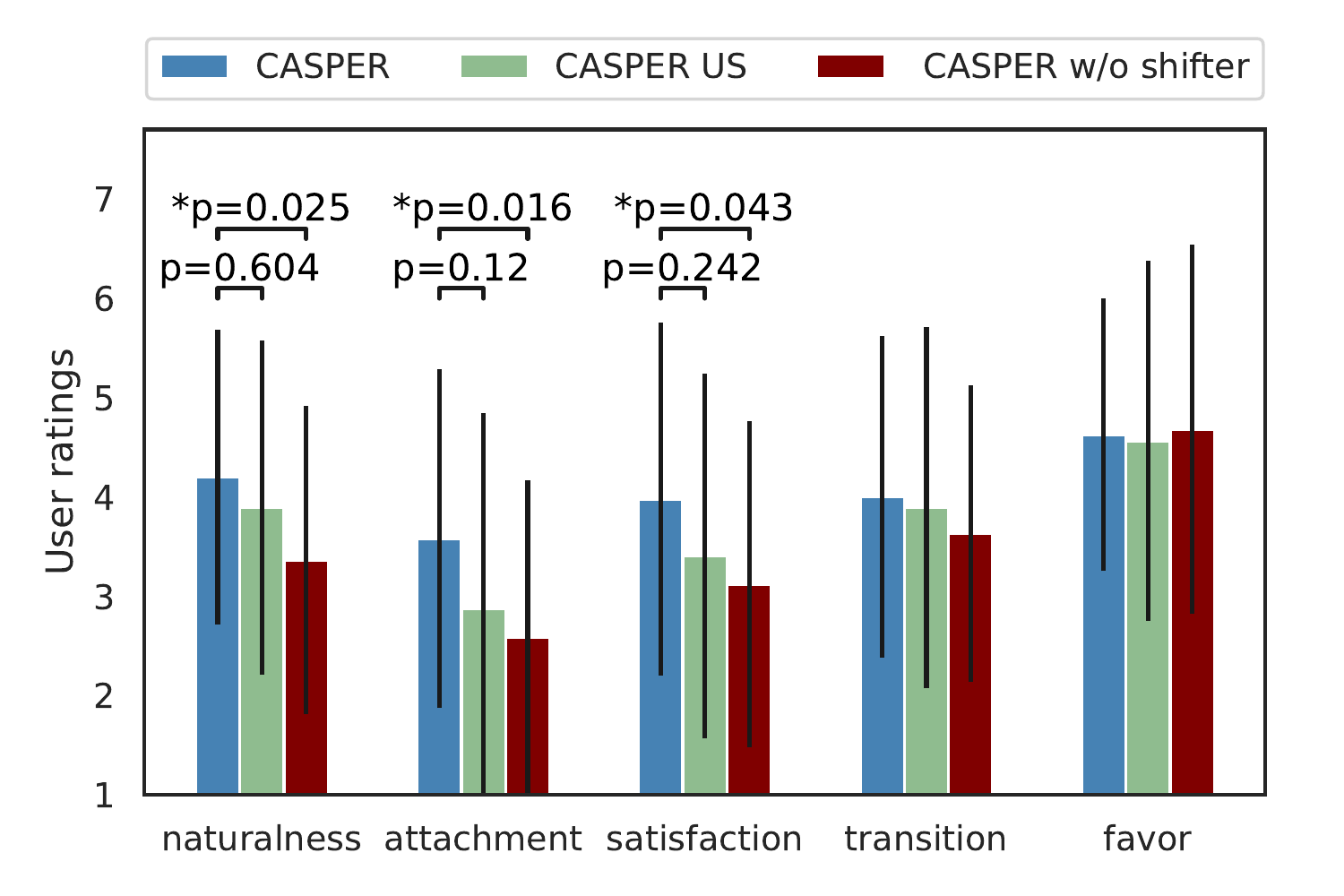}
  \caption{
    \textbf{Comparison of user ratings for each question among CASPER, CASPER US, and CASPER w/o shifter}.\label{fig:ablation}
    Mean and standard deviation of ratings for each item are shown as histograms and bars, respectively.}
\end{figure}

\section{Experiment\label{sec:experiment}}
We conducted experiments to investigate our research questions:
(RQ~1) whether selector activates the model at the appropriate time and to make the dialog flow seamless and
(RQ~2) whether shifter generates intermediate utterances to bridge the gap between chat- and task- oriented dialog.

\subsection{Procedure}
In the experiments, we conducted objective evaluations and user studies.
In the objective evaluations, we evaluated the fluency of the systems' responses by comparing perplexity with CASPER and baseline.
In the user studies, participants were asked to interact with a dialog system and were then asked to rate the items listed in Table \ref{table:eval_items} on a 7-point scale.
We validated our overall CASPER design by comparing these user ratings between CASPER and baseline (RQ~1, and 2).
We also evaluated the effectiveness of shifter by comparing user ratings among the original CASPER with the ablation models (RQ~2).

After the conversation, participants answered a questionnaire containing all the questions listed in Table~\ref{table:eval_items} on a 7-point scale.
In particular, transition and naturalness are the items that are aimed to validate the research questions (RQ~1, and 2).
We expect the following results:
(1) detecting the appropriate timing of domain transitions by selector reduces the forcefulness and increases the transition score and
(2) appropriate intermediate utterance by shifter increases the naturalness score.

\subsection{Experimental Setup\label{subsec: setup}}
\noindent\textbf{Dataset. }
All dialog samples were collected from Twitter, and the total number of context-reply pairs used for the entire training of CASPER was 17.8M pairs. The annotation of domain follows the annotation of Twitter API v2. We used three types of domains: TV programs, musicians, and video games, and prepared a {\ltwo} and {\lthree} for each domain.

\noindent\textbf{Implementation of CASPER.}
For the implementation of chatter and shifter, we used BERT2BERT \cite{bert2bert}, which is an encoder-decoder model using BERT checkpoints. The document classification model for the selector was based on BERT \cite{bert}. Both were implemented by fine-tuning \textit{cl-tohoku/bert-base-japanese-whole-word-masking}, a pre-trained checkpoints of BERT in Japanese.

We constructed {\lthree}, a simple rule-based model, to recommend products through keyword matching and co-reference resolution by collecting words that occur frequently in tweets containing the name of an entity but are not likely to be included in tweets of other entities. We chose such a simple model to avoid differences in experimental results due to the behavior of the recommendation algorithm.

\noindent\textbf{Baseline and ablation.}
The baseline dialog system corresponds to a single-model dialog system. That is, it performs all the roles of the models in CASPER alone. Its model is implemented as the same BERT2BERT model as {\lone}. The training procedure is as follows:
We first trained the model in the same way as {\lone}. Then, we performed transfer learning with all the datasets used for {\ltwo} models to learn domain-specific dialog. Lastly,
the baseline learned recommendation dialogs across the three domains in an end-to-end manner as in a previous study \cite{fb_end2end}.

We also conducted an ablation study to verify the following two points about shifter: (1) whether shifters contribute smooth topic transition and (2) whether it is appropriate to train shifter separately for each domain. To address these points, we prepared the following two ablation systems for comparison with the original CASPER. (1) CASPER without shifter (CASPER w/o shifter) is a model with only chatter and performer (without shifter), corresponding to bi-model systems.(2) CASPER unified shifter (CASPER US) is a model trained on a single shifter without separating the corpus and model for each domain.

\noindent\textbf{Conversation.}
All conversations were conducted in Japanese.
For each model, we crowdsourced 50 different participants and had them interact with the system and the user on a web site of our own design. The conversation ended when the participant accepted the product recommendation or when 40 time steps elapsed. The acceptance of a product recommendation was determined by a pop-up on the web system when the participant refers to the product positively. After the conversation, the participants answered a questionnaire containing all the questions listed in Table. \ref{table:eval_items} on a 7-point scale.

\subsection{Results}
\noindent\textbf{Objective evaluation.}
The results of the comparison of perplexity and naturalness are shown in Table \ref{table:ppl}. The perplexity of CASPER was calculated from the average perplexities from chatter and all shifters. CASPER was slightly lower than the baseline with a difference of 0.22. In the user assessment of naturalness, CASPER exceeded the baseline by a difference of 0.79. The difference was statistically significant (Student's t-test).

\noindent\textbf{Comparison with baseline.}
The experimental results for the comparison of user evaluation between CASPER and baseline are shown in Fig. \ref{fig:base_casper}.
In addition to the aforementioned naturalness, in two other items, attachment and transition, CASPER scored significantly higher than the baseline. For satisfaction, CASPER scored higher than baseline by a marginally significant difference. For favor, there was no significant difference between CASPER and baseline.

\noindent\textbf{Comparison with ablated systems.}
The experimental results for CASPER and ablated systems are shown in Fig.~\ref{fig:ablation}.
The one-way ANOVA showed that there were significant differences for naturalness, attachment, and satisfaction ($p < .05$).
As post-hoc analysis, we conducted multiple comparisons for using Tukey's test.
Compared with CASPER US, the original CASPER was slightly higher in attachment and satisfaction by 0.71 and 0.58, respectively, but the differences were not significant ($p > .1$).
Compared with CASPER w/o shifter, the original CASPER was significantly higher in naturalness, attachment, and satisfaction by 0.86, 0.99, and 0.86, respectively ($p < .05$).
\section{Discussion}
\begin{table}[t]
  \centering
  \begin{tabular}{rll}
    \hline
    Speaker   & text                                    \\
    \hline\hline
    user&What should I call you?              \\
    chatter&Call me whatever you want.           \\
    user&OK, I'll call you Mr. Anonymous.     \\
    chatter&Yes, sir! What would you like me to call you?\\
    user&Please call me xx. (authors anonymized)\\
    …&…\\
    chatter&What kind of hobbies do you like?  \\
    user&I like to travel.                   \\
    shifter&Traveling sounds nice! I want to travel too!\\
    &I want to go on a anime pilgrimage to\\
    &the locations for \textit{Laid-Back Camp}!\\
    …&…\\
    performer&How about \textit{Laid-Back Camp}?      \\
    system&task success: \textit{Laid-Back Camp}   \\
    \hline
  \end{tabular}
  \caption{
    \textbf{Interaction between participant and CASPER during experiment}\label{table:ours_example}
  }
\end{table}

\begin{table}[t]
  \centering
  \begin{tabular}{rll}
    \hline
    Speaker  & text                                     \\
    \hline\hline
    user     & What are you addicted to these days?     \\
    baseline & An anime called \textit{Tokyo Revengers} is interesting. \\
    user     & I see. Anime. I sometimes read comics. \\
    baseline & I recommend \textit{Tokyo Revengers}! \\
    …&…\\
    baseline & I like the songs of \textit{BUMP OF CHICKEN}!\\
    &I recommend them!\\
    user     & What are some famous songs?              \\
    baseline & They're all famous!    \\
    …&…\\
    system   & task success: \textit{BUMP OF CHICKEN}            \\
    \hline
  \end{tabular}
  \caption{
    \textbf{Interaction between participant and baseline during experiment}
    \label{table:base_example}
  }
\end{table}
\subsection{Comparison with Baseline}
The dialog of CASPER was more natural (naturalness), less forced (transition), and more satisfying (satisfaction) compared to baseline, which suggests that CASPER can bridge the gap between chat- and task-oriented dialog better than a single-model system.
One possible reason for this is that the recommendation dialog was carefully developed over a longer time step than that in the baseline. In practice, the median and mean time steps of the first recommendation were 5 and 9.65 for the baseline, and 13 and 13.7 for CASPER.

To verify the interaction with the two dialog systems in more detail, we show the examples of CASPER and baseline dialogs in Tables \ref{table:ours_example} and \ref{table:base_example}, respectively. The baseline showed a dialog to accomplish a task by responding with praise and recommendations of products. CASPER, on the other hand, waited for the appropriate moment to make a natural topic transition by continuing a task-unrelated dialog for a while and shifted the topics naturally at specific times while maintaining consistency in questions and answers.

The quantitative evaluation of naturalness and transition and the qualitative evaluation using dialogue examples indicated that the selector appropriately selected the topic transition destination and timing, and shifter was able to make a natural transition of the topic to the target domain while maintaining consistency between the question and the answer.
In other words, CASPER was able to approach two issues that have not been sufficiently examined in prior studies: the seamless connection between chat- and task-oriented dialogue and the appropriate timing of switching between the two.
CASPER also outperformed the baseline in attachment and satisfaction, indicating that it is more suitable as an artificial-intelligence assistant at home.
Unlike store assistant robots against which people are likely to be prepared for task-oriented dialogue, the policy of performing a task only when it is natural would be suitable for domestic use of dialogue systems.

\subsection{Comparison with Ablation Systems}
The original CASPER provided more natural responses and higher dialog satisfaction compared to CASPER w/o shifter, which corresponds to bi-model systems. This indicates that the topic transition of shifter not only benefits the designer by guiding the topic to the task but also benefits the user by making the interaction natural and satisfying. One of the reasons is that shifter's moderate topic provision reduced the dullness of the chatter's generic-response dialog.

We expected that the original CASPER to score higher than CASPER US, which employs a unified shifter for all the target task domains,
because the unified model would become overly generalized due to learning with responses from various domains.
However, although CASPER US scored slightly lower in attachment and satisfaction than CASPER, we could not find significant differences.
This results indicate that the unified shifter is capable of managing three domains, and a new question of how many tasks it can manage has remained for future work.
We would also like to further investigate the controllability of topics by changing the $\alpha$ parameters of selector (Sec. \ref{ss:selector}).

\section{Conclusion}
We proposed CASPER, a dialog system that bridges the gap between task- and non-task-oriented dialog using the domain shift.
Shifter and selector enables a natural and seamless flow of dialog from open-domain chat to task-oriented conversation.
In a user study, we showed that CASPER gave a better impression in terms of naturalness of response, lack of forced topic switching, and satisfaction compared with a baseline.
We also discussed how shifter and selector naturally introduced task-oriented dialog by conducting a qualitative evaluation of the conversation. Selector detected opportunities for natural topic transitions, and shifter seamlessly changed topic, maintaining consistency in questions and answers.
\section*{Acknowledgements}
This work was supported by JST CREST Grant Number JPMJCR19A1, Japan.
\bibliographystyle{named}
\bibliography{main}

\begin{thebibliography}{}

\bibitem[\protect\citeauthoryear{Akasaki and Kaji}{2017}]{chat_detection}
Satoshi Akasaki and Nobuhiro Kaji.
\newblock Chat detection in an intelligent assistant: Combining task-oriented
  and non-task-oriented spoken dialogue systems.
\newblock In {\em Proceedings of the 55th Annual Meeting of the Association for
  Computational Linguistics (Volume 1: Long Papers)}, pages 1308--1319, 2017.

\bibitem[\protect\citeauthoryear{Bordes \bgroup \em et al.\egroup
  }{2016}]{fb_end2end}
Antoine Bordes, Y-Lan Boureau, and Jason Weston.
\newblock Learning end-to-end goal-oriented dialog.
\newblock {\em arXiv preprint arXiv:1605.07683}, 2016.

\bibitem[\protect\citeauthoryear{Chen \bgroup \em et al.\egroup }{2019}]{kbrd}
Qibin Chen, Junyang Lin, Yichang Zhang, Ming Ding, Yukuo Cen, Hongxia Yang, and
  Jie Tang.
\newblock Towards knowledge-based recommender dialog system.
\newblock In {\em Proceedings of the 2019 Conference on Empirical Methods in
  Natural Language Processing and the 9th International Joint Conference on
  Natural Language Processing (EMNLP-IJCNLP)}, pages 1803--1813, Hong Kong,
  China, November 2019. Association for Computational Linguistics.

\bibitem[\protect\citeauthoryear{Devlin \bgroup \em et al.\egroup
  }{2019}]{bert}
Jacob Devlin, Ming-Wei Chang, Kenton Lee, and Kristina Toutanova.
\newblock Bert: Pre-training of deep bidirectional transformers for language
  understanding.
\newblock In {\em NAACL-HLT (1)}, pages 4171--4186, 2019.

\bibitem[\protect\citeauthoryear{Liu \bgroup \em et al.\egroup
  }{2018}]{less_generic_response}
Yahui Liu, Wei Bi, Jun Gao, Xiaojiang Liu, Jian Yao, and Shuming Shi.
\newblock Towards less generic responses in neural conversation models: A
  statistical re-weighting method.
\newblock In {\em Proceedings of the 2018 conference on empirical methods in
  natural language processing}, pages 2769--2774, 2018.

\bibitem[\protect\citeauthoryear{Liu \bgroup \em et al.\egroup }{2019}]{akgcm}
Zhibin Liu, Zheng-Yu Niu, Hua Wu, and Haifeng Wang.
\newblock Knowledge aware conversation generation with explainable reasoning
  over augmented graphs.
\newblock In {\em Proceedings of the 2019 Conference on Empirical Methods in
  Natural Language Processing and the 9th International Joint Conference on
  Natural Language Processing (EMNLP-IJCNLP)}, pages 1782--1792, 2019.

\bibitem[\protect\citeauthoryear{Lubis \bgroup \em et al.\egroup
  }{2019}]{emo_hred}
Nurul Lubis, Sakriani Sakti, Koichiro Yoshino, and Satoshi Nakamura.
\newblock Positive emotion elicitation in chat-based dialogue systems.
\newblock {\em IEEE/ACM Transactions on Audio, Speech, and Language
  Processing}, 27(4):866--877, 2019.

\bibitem[\protect\citeauthoryear{Ma \bgroup \em et al.\egroup
  }{2021}]{cr_walker}
Wenchang Ma, Ryuichi Takanobu, and Minlie Huang.
\newblock Cr-walker: Tree-structured graph reasoning and dialog acts for
  conversational recommendation.
\newblock In {\em Proceedings of the 2021 Conference on Empirical Methods in
  Natural Language Processing}, pages 1839--1851, 2021.

\bibitem[\protect\citeauthoryear{Nakano \bgroup \em et al.\egroup
  }{2006}]{multi_expert_model}
Mikio Nakano, Atsushi Hoshino, Johane Takeuchi, Yuji Hasegawa, Toyotaka Torii,
  Kazuhiro Nakadai, Kazuhiko Kato, and Hiroshi Tsujino.
\newblock A robot that can engage in both task-oriented and non-task-oriented
  dialogues.
\newblock In {\em 2006 6th IEEE-RAS International Conference on Humanoid
  Robots}, pages 404--411. IEEE, 2006.

\bibitem[\protect\citeauthoryear{Rothe \bgroup \em et al.\egroup
  }{2020}]{bert2bert}
Sascha Rothe, Shashi Narayan, and Aliaksei Severyn.
\newblock Leveraging pre-trained checkpoints for sequence generation tasks.
\newblock {\em Transactions of the Association for Computational Linguistics},
  8:264--280, 2020.

\bibitem[\protect\citeauthoryear{Sun and Zhang}{2018}]{crm}
Yueming Sun and Yi~Zhang.
\newblock Conversational recommender system.
\newblock In {\em The 41st international acm sigir conference on research \&
  development in information retrieval}, pages 235--244, 2018.

\bibitem[\protect\citeauthoryear{Vinyals and Le}{2015}]{seq2seq}
Oriol Vinyals and Quoc Le.
\newblock A neural conversational model.
\newblock In {\em Deep Learning Workshop 2015}, 2015.

\bibitem[\protect\citeauthoryear{Zeng \bgroup \em et al.\egroup
  }{2021}]{adversarial_chat_detection}
Zhiyuan Zeng, Hong Xu, Keqing He, Yuanmeng Yan, Sihong Liu, Zijun Liu, and
  Weiran Xu.
\newblock Adversarial generative distance-based classifier for robust
  out-of-domain detection.
\newblock In {\em ICASSP 2021-2021 IEEE International Conference on Acoustics,
  Speech and Signal Processing (ICASSP)}, pages 7658--7662. IEEE, 2021.

\bibitem[\protect\citeauthoryear{Zhao \bgroup \em et al.\egroup
  }{2017}]{ner_seq2seq}
Tiancheng Zhao, Allen Lu, Kyusong Lee, and Maxine Eskenazi.
\newblock Generative encoder-decoder models for task-oriented spoken dialog
  systems with chatting capability.
\newblock In {\em Proceedings of the 18th Annual SIGdial Meeting on Discourse
  and Dialogue}, pages 27--36, 2017.

\end{thebibliography}
\end{document}